\documentclass[conference, 12pt]{IEEEtran}
\IEEEoverridecommandlockouts
\usepackage{cite}
\usepackage{amsmath,amssymb,amsfonts}
\usepackage{algorithmic}
\usepackage{tabularray}
\usepackage{graphicx}
\usepackage{textcomp}
\usepackage{fancyhdr}
\usepackage{xcolor}
\usepackage{float}
\usepackage[T1]{fontenc}
\def\BibTeX{{\rm B\kern-.05em{\sc i\kern-.025em b}\kern-.08em
    T\kern-.1667em\lower.7ex\hbox{E}\kern-.125emX}}
\usepackage{fancyhdr}

\fancypagestyle{IEEEtitlepagestyle}{
  \fancyhf{} 
   
  \setlength{\headheight}{15.31181pt}
  \fancyhead[R]{\raisebox{-1ex}{Duke Robotics Club | \thepage}}
}
\begin{document}

\title{\textbf{Oogway: Designing, Implementing, and Testing an AUV for  Robosub 2023} 
}

\author{Will Denton, Lilly Chiavetta, Michael Bryant, Vedarsh Shah, Rico Zhu, Ricky Weerts,\\Phillip Xue, Vincent Chen, Hung Le, Maxwell Lin, Austin Camacho, Drew Council,\\Ethan Horowitz, Jackie Ong, Morgan Chu, Alex Pool}

\maketitle

\begin{abstract}
\normalsize
The Duke Robotics Club is proud to present our robot for the 2023 RoboSub Competition: Oogway. Oogway marks one of the largest design overhauls in club history. Beyond a revamped formfactor, some of Oogway's notable features include all-new computer vision software, advanced sonar integration, novel acoustics hardware processing, and upgraded stereoscopic cameras.

Oogway was built on the principle of independent, well-integrated, and reliable subsystems. Individual components and subsystems were tested and designed separately. Oogway's most advanced capabilities are a result of the tight integration between these subsystems. Such examples include sonar-assisted computer vision algorithms and robot-agnostic controls configured in part through the robot's 3D model.

The success of constructing and testing Oogway in under 2 year's time can be attributed to 20+ contributing club members, supporters within Duke's Pratt School of Engineering, and outside sponsors. 
\end{abstract}

\section{Competition Goals}
In previous years of the RoboSub competition we prioritized modality and flexibility over reliability. While our previous designs allowed us to complete more tasks, our runs were inconsistent and unreliable. Since we only have three total scoring runs we need to ensure reliability and the completion of all attempted tasks in each run.

This year, our strategic vision consists of \textit{core tasks} which the robot is programmed to perform every run with one \textit{alternate task} which may be completed at the end of each successful run if necessary. 

RoboSub's design goals focus on four fundamental principles: movement, vision, manipulation, and acoustic tracking. To accomplish our strategic vision we chose to focus on 3 of 4 design goals: movement, vision, and acoustic tracking. This subset of goals aligns with our team's strengths in controls, computer vision (CV), and electronics. Our focus on these methods enables us to perform the gate, buoy, and octagon tasks (our \textit{core tasks}) reliably, while still maintaining the flexibility to attempt the torpedoes task (our \textit{alternate task}) if needed. 

\subsection{Strategic Vision}
The team's strategy this year focuses on completing a subsection of tasks while having fail-safes in place to ensure a successful run. We designed Oogway for this year's competition to ensure that we can always complete \textit{core tasks}, and in the event of a failure, we have multiple methods to complete each task in the form of fallback methods. 

Movement is essential for every task so we thoroughly tuned our cascaded PID control system for optimal movement in 6 directions. For both the CV and acoustics systems, we rewrote the codebase, adding a new stereovision camera and additional high resolution hydrophones. We mounted a new sonar sensor to obtain real-time depth information of objects in front of the robot. These new components allow us to achieve our strategic vision through multiple methods of completing the gate, buoy, and octagon tasks.

\subsection{Destination - Gate}
Since the gate task is required to start a run, we used multiple subsystems to ensure that we could reliably move through the gate. With our redesigned CV system, Oogway is able to identify the gate from any angle and get an estimated position of the poles and center. We implemented a sonar tracking system to identify the gate with minimal position error. Given these two fallback methods we will elect to do the coin flip for the random starting heading since Oogway can reliably identify the gate. We will always pass through the Abydos side of the gate to have the same competition run. We are confidant we can complete the 720$^{\circ}$ yaw style tasks with our tuned mechanical control system.

\subsection{Start Dialing - Buoy}
The buoy task is another core component of our competition strategy. After we complete the spin, we use our sonar-CV system to identify and get the position of the buoy. Since we do not have a downward facing camera on Oogway, we do not use the path markers to track the buoy. Instead, we use the sonar which can scan objects up to 30m away. Once in position in front of the buoy, we hit both corresponding symbols and then back up. This task is one of our strongest tasks as we designed Oogway to have precise CV and mechanical control.

\subsection{Engaging Chevrons - Octagon}
The final component of our core tasks is surfacing within the octagon. Since robot actuators are still in development, we opt to not move the chevrons. After completing the buoy task we activate our acoustic system, listen for the acoustic signal, move slowly in the direction of the signal, and surface in the octagon. Should we encounter issues with the acoustics, the sonar is used as a fallback to locate the top of the octagon and surface.

In preparation for the task of locating the octagon, developing a reliable acoustics system was a large component of our design. Acoustic hydrophones listen for the ping of the underwater pinger and calculate the direction of the source, providing directional guidance for the robot.

\subsection{Goa’uld Attack - Torpedoes}
Even though we have minimally tested the torpedoes task, if we need the additional points at competition we believe that we can complete the torpedoes task after surfacing in the octagon. Our new CV algorithm is fully capable of identifying the torpedo circle. However, we are unsure of the capabilities of the torpedo launcher, given our focus on vision and acoustic tracking. With additional development time at competition, we believe we can accomplish this task as a risky end to our run.

\subsection{Location - Bins}
We are not planning on attempting the bins task since Oogway does not have actuators or a downward facing camera. Given our team's time and member constraints, we decided to forgo this task until next year's competition.

\subsection{Trade Offs Between Complexity and Reliability.}
Given last year's competition experience, we realized that many of our systems were too complex. We attempted to do every task which made it impossible to thoroughly test each component in the year-long span of creating a new robot. We designed Oogway such that each component is separate, allowing for thorough testing, while being robust enough to complete the tasks successfully. By narrowing our focus to 3 out of the 5 tasks during the design process for this year's competition, we allocated more testing time for each task to ensure that our complex integration can work reliably. 

\section{Design Strategy}

\subsection{Mechanical Design}
\begin{figure}[h!]
\centering
\includegraphics[width=69mm]{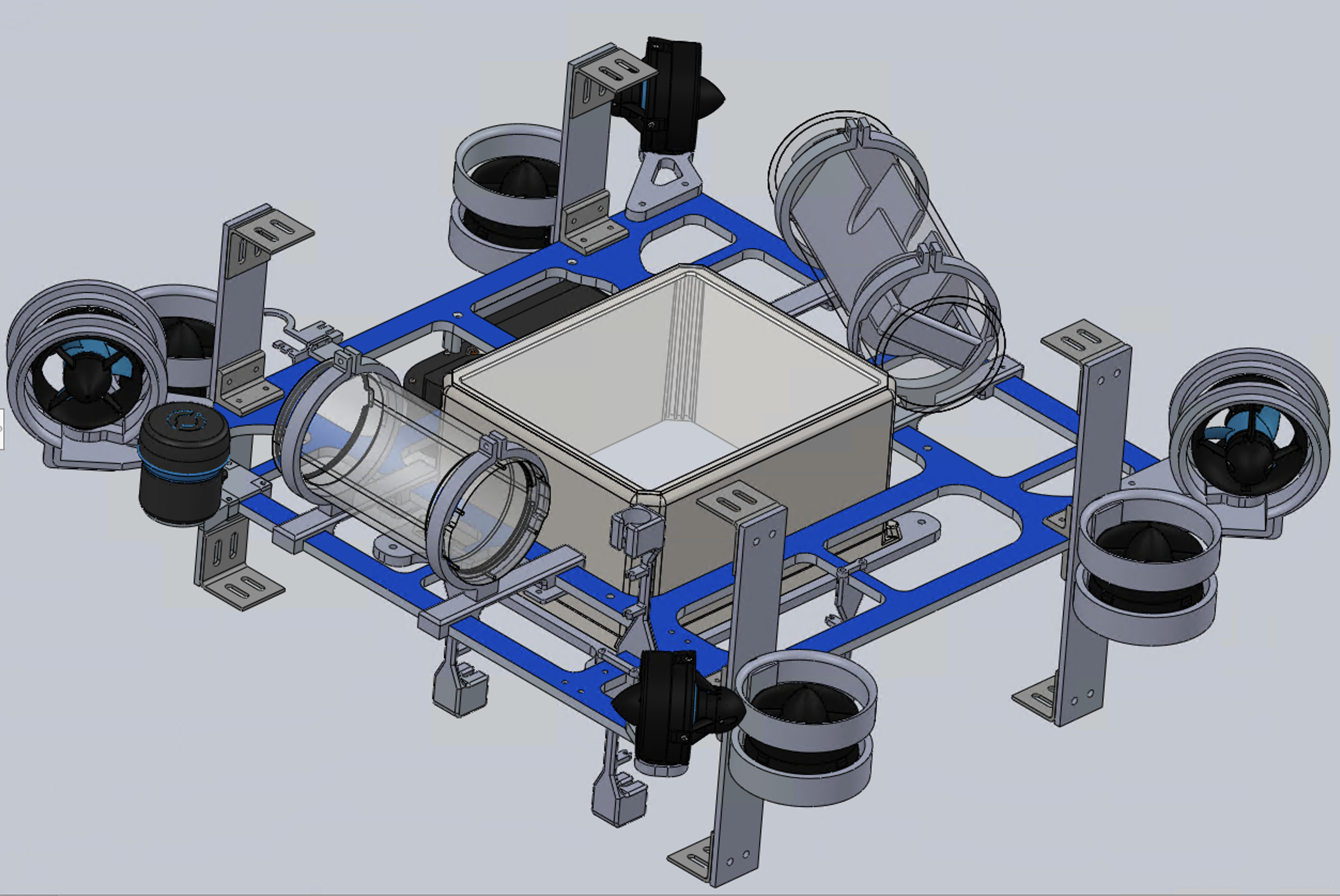}
\caption{Full SOLIDWORKS Assembly of Oogway}
\label{fig:method}
\end{figure}

The mechanical design of Oogway employs a wide structure to position the thrusters further away from the center of mass. This allows them to produce more torque for actions such as spinning. 

Oogway is also designed with adaptability in mind. The frame has mounting rails built directly into it, offering enhanced modularity. Every component can be re-positioned easily, allowing optimization for factors such as buoyancy. Additionally, this modularity supports the independent integration of the CV cameras and sonar subsystems.

\begin{figure}[h!]
\centering
\includegraphics[width=69mm]{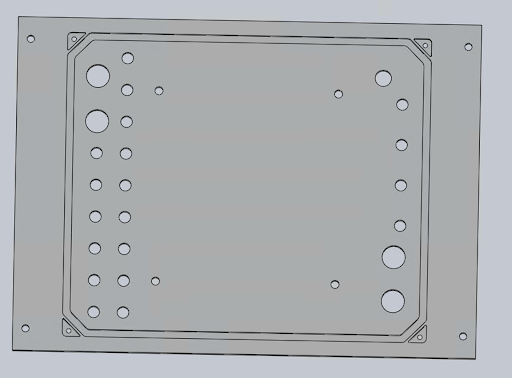}
\caption{Aluminium base plate built for adaptability}
\label{fig:method}
\end{figure}

Additionally, the aluminium plate at the bottom of the stack was built to incorporate many sensors and electrical attachments. It was designed with more holes than needed so it can be adaptable to additional sensors that need to be attached. After construction, the plate was self-anodized by the team to prevent corrosion.

\subsection{Electrical Design}
\subsubsection{Electronics Stack Architecture}

One of our design objects around Oogway's electrical stack is based on the idea of supporting a variable number of systems while making small changes easy. Our previous robots integrated the battery within the rest of the electrical hardware. Oogway iterates on that approach by adding a separate capsule that isolates the battery from the stack. This streamlines the common task of replacing the battery and reduces the risk of damaging other seldom-accessed components. It also provides physical protection, should the battery have an internal problem. 

We added a separate capsule to create a new interchangeable camera system. This camera system allows us to use multiple cameras without removing components from the stack. Our new cameras have onboard processing for computer vision algorithms to be executed, taking load off the main robot computer. This also helps distribute heat within the robot as it separates one of its biggest thermal producers to an area where it can cool more effectively.

The stack's optimizations extend beyond isolating components. Our club's electrical and mechanical subteams collaborated to reduce cable lengths and increase component accessibility. For example, our powerbus now supports a modular number of components with different voltage levels which was necessary to integrate different sonars and cameras. These advances help maintain and improve the electrical performance of components, such as reducing impedances in analog signal wires.

\begin{figure}[h!]
\centering
\includegraphics[width=69mm]{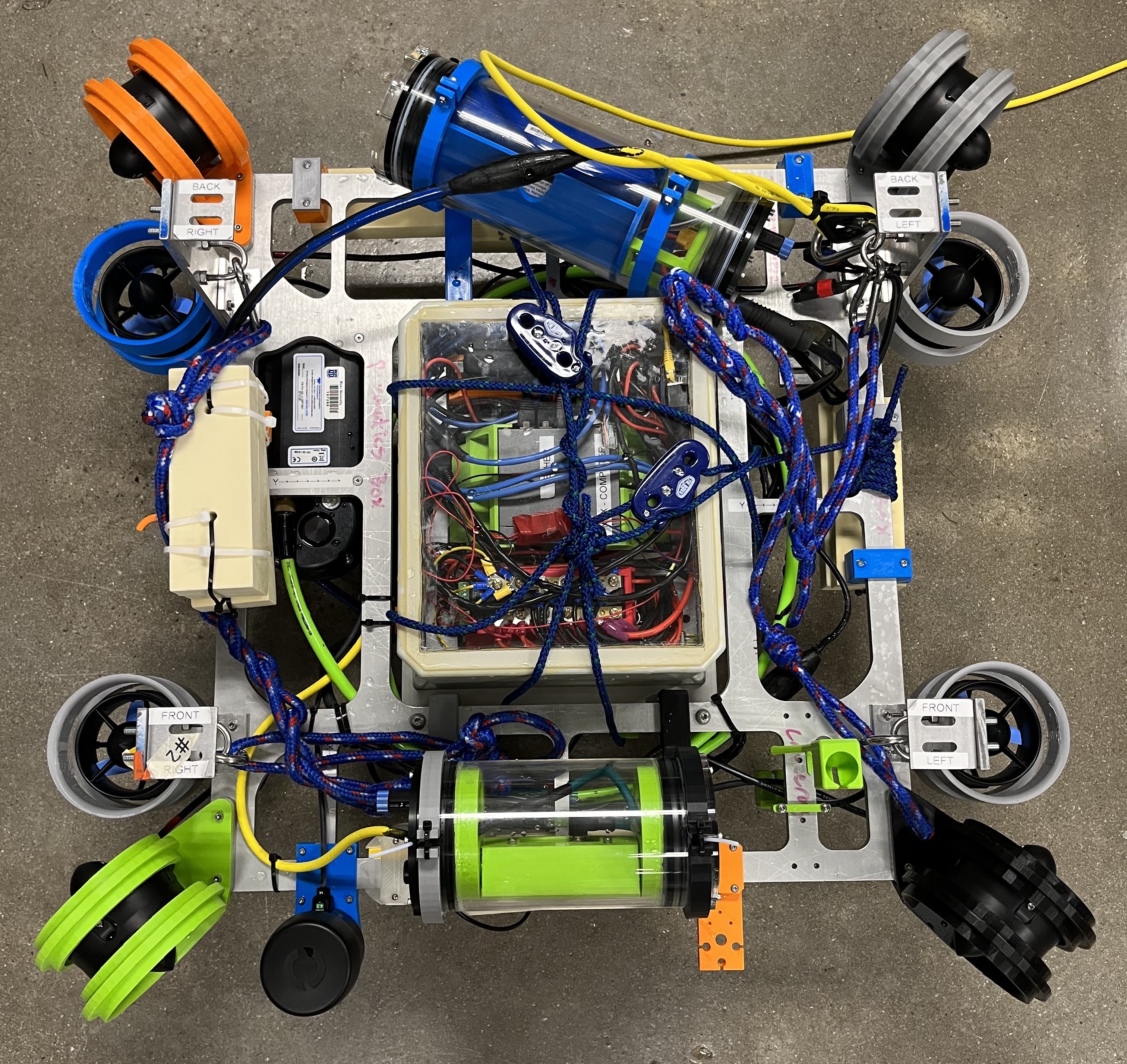}
\caption{The stack sits prominently at the center of Oogway. This is also a design consideration as the IMU, located inside, is mounted at the center of mass of the robot.}
\label{fig:method}
\end{figure}

The bottom of the stack contains an array of plugs that can be adapted to use any size subcon cable. This enables the addition of external components and even allows some sensors, such as pressure sensors, to sit on the boundary of the stack.

\subsubsection{Acoustics}
Our acoustics stack on Oogway is a custom built subsystem designed with the specific objective of locating the underwater pinger in the octagon. A critical component of this system includes hand-populated SMT boards that offer a compact formfactor and signal impedence matching. The data from these boards is pipelined directly into an onboard DAQ where the in-house software calculates and reports the azimuth of the pinger to a ROS topic.

The electrical hardware of the acoustics subsytem is the accumulation of years of testing and development. We formalized findings from prior perfboard wirings and variable test circuits as custom PCBs. Oogway sports 8 hydrophones, each with its own filters and amplifiers. All of this custom hardware is mounted on two PCBs, which combined is smaller than the area of a credit card.

Though originally designed with low-pass filters, when mounted to Oogway, low frequency noise from the thrusters and other inductive loads deteriorated the signal-to-noise ratio (SNR). To surmount this, we devised a bandpass filter (BPF). Remarkably, this did not require any modification to the existing circuitry as a backpack was designed to attach directly to the already-made boards.

\begin{figure}[h!]
\centering
\includegraphics[width=69mm]{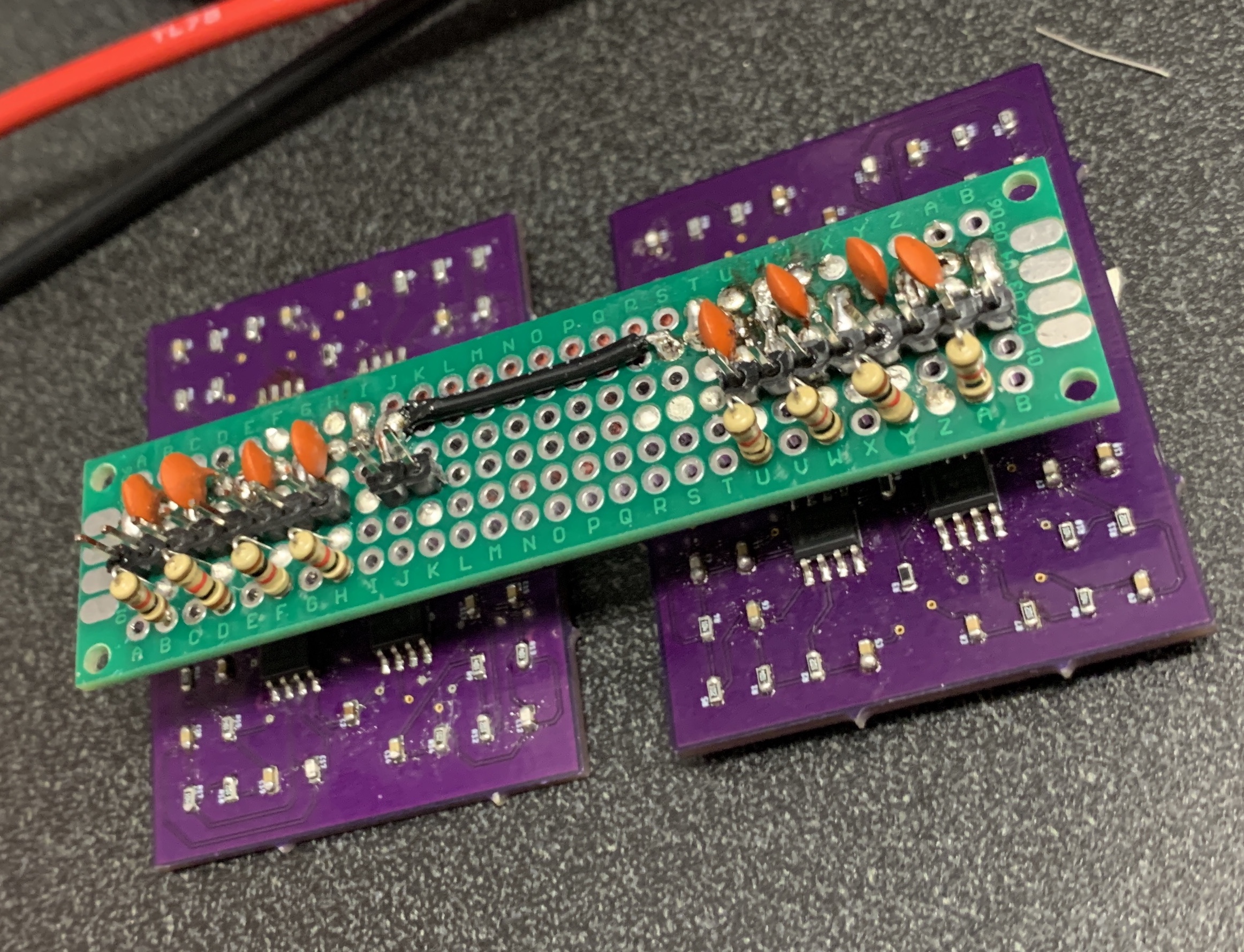}
\caption{Acoustics Filter and Amplifier Boards with BPF Backpack}
\label{fig:method}
\end{figure}

With the improved filters, we constructed a testing setup wherein the gain of the amplifiers could be changed in real time. Through a series of tests in a pool, a gain on the order of magnitude of $10 ^ 1 V/V$ provided visual indication of the ping in the raw data, even from tens of meters away. 

\begin{figure}[h!]
\centering
\includegraphics[width=69mm]{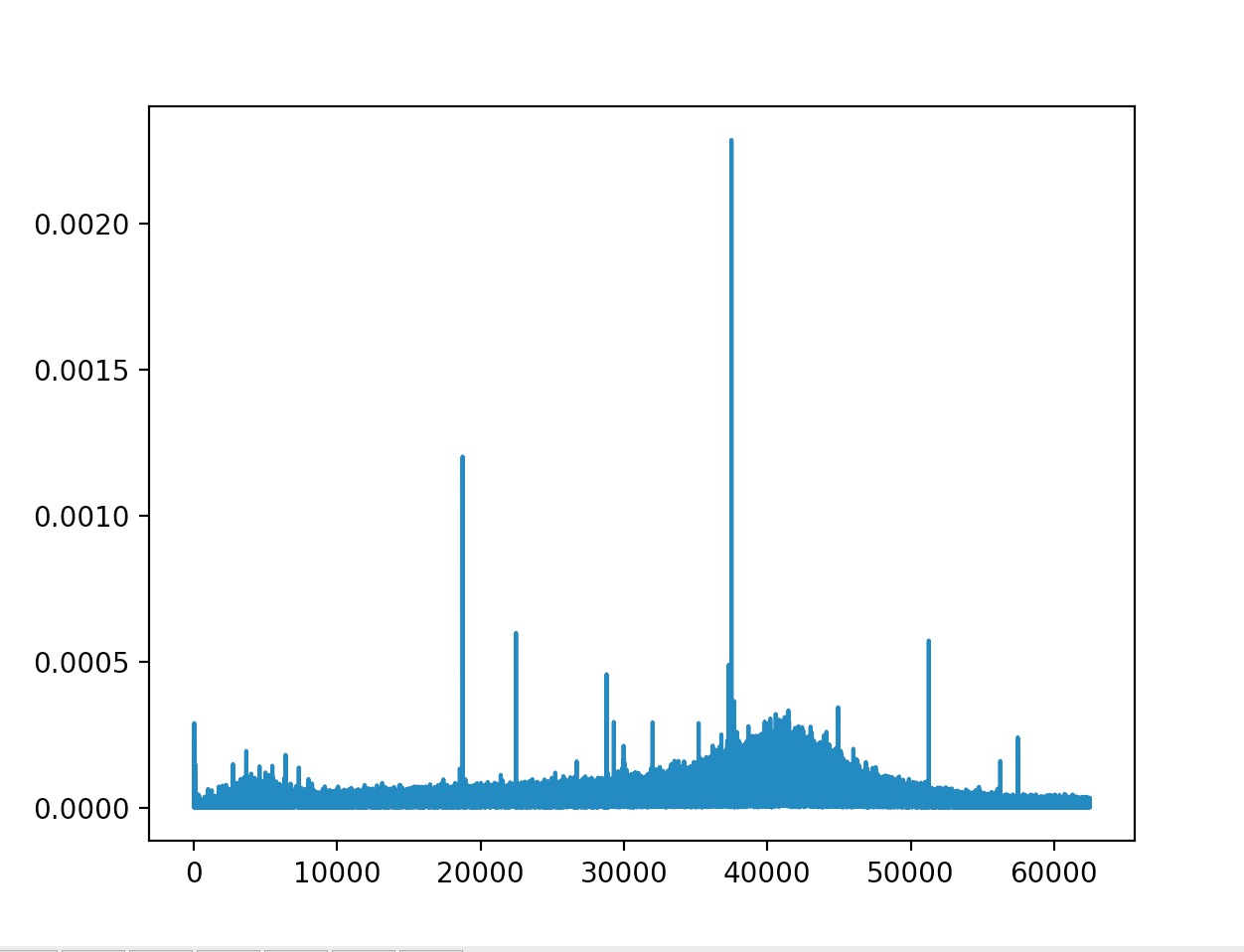}
\caption{Magnitude of Frequency Response of Filters. Bandpass 'hump' visible at $\sim$40kHz contains ping. Attenuated $\sim$18kHz noise also visible.}
\label{fig:method}
\end{figure}

The acoustics software is based on a 4-dimensional time-difference-of-arrival optimization problem solved using gradient descent. We optimized this calculation by a computationally cheap time-of-arrival-based guessing server.\cite{b1}

A software Butterworth filter feeds into a correlation to determine relative phase differences between channels. The algorithm chooses a window with relatively small variance in phase differences. By taking pairwise comparisons with four hydrophones and with Euclidean geometry, the algorithm can set up the gradient descent problem. The hydrophones are located within 1/2 wavelength of each other to mitigate aliasing.

Oogway's other four hydrophones are dedicated to providing a guess aimed at octant-level accuracy. These hydrophones are located on opposite sides of the robot and use absolute timings to determine the guessed pinger location for the octagon. This guess greatly increases our probability of determining the global optimization in pinger location.

This new acoustics subsystem provides a reliable way to accurately locate all the acoustic pingers for completing competition tasks.

\subsection{Software Design}

\subsubsection{Computer Vision}
One major focus of our CV subteam this year was improving existing tools for a smoother and more efficient procedure for testing.

During testing and footage collection for last year’s competition, we could only extract footage once the robot was out of the water. We were unable to ensure the quality of our data, leading to vast amounts of useless, low-quality data. While developing Oogway we created a Graphical User Interface (GUI) which allows us to view live feed from the underwater robot to make immediate changes as necessary. The GUI also includes different widgets that allow users to record bag footage, launch/track ROS nodes and topics, view live bounding boxes, and control/track camera status.

\begin{figure}[h!]
\centering
\includegraphics[width=69mm]{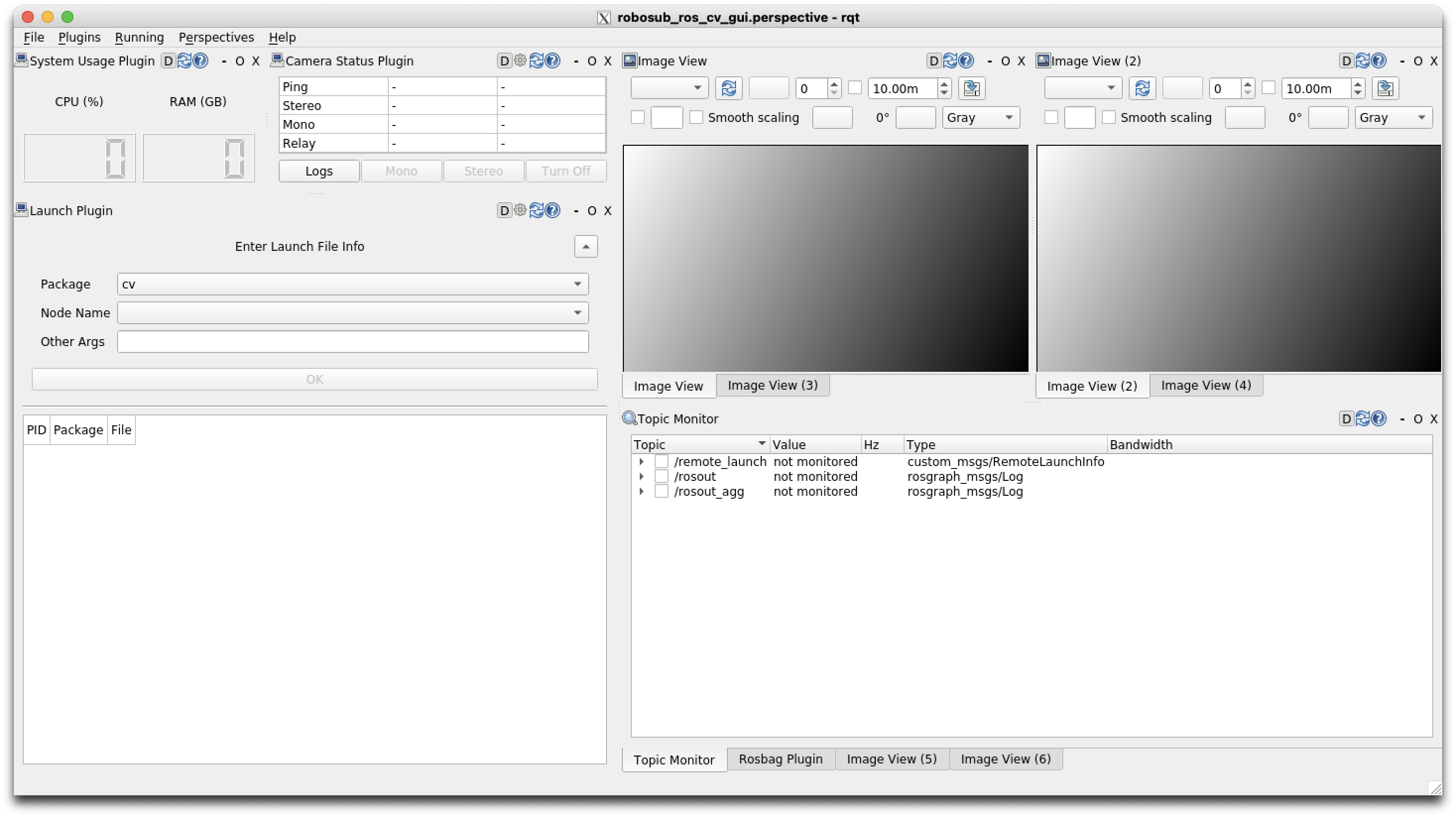}
\caption{CV GUI}
\label{fig:method}
\end{figure}

A second goal of our CV team was to research and decide on a new CV object detection model. We previously used a YOLOv4-tiny model for object detection; however, we had many extraneous detections. To ensure reliability of the CV subsystem, it was necessary to have a higher detection rate and accuracy. After researching and comparing the performance of various CV models on our datasets, we upgraded our model to the YOLOv7-tiny.\cite{b2} We found this model could process camera data at 10Hz with over a 95\% accuracy.

Our real images were captured in a competition-like environment with different glyph orientations and camera angles.

\begin{figure}[h!]
\centering
\includegraphics[width=69mm]{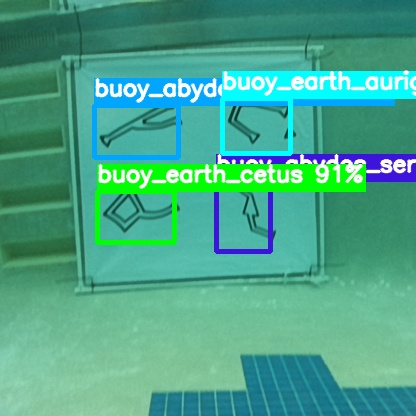}
\caption{Buoy Successfully Detected, Indicated with Bounding Boxes}
\label{fig:method}
\end{figure}

\subsubsection{Sonar}
Another focus of our CV team this year was a holistic approach toward depth tracking. Last year, we used a dual mono-camera based stereo vision system to estimate the depth of the objects we detected. We noted two key limitations which upon evaluation lead us to redesign our system for gathering depth data; (1) the calibration process to synchronize the two cameras was time consuming and prone to error; (2) even after calibration, our depth estimation accuracy was still poor and had high variance in performance between runs, especially in response to changes in lighting conditions.

With more testing, we quickly noted that stereo vision was not a reliable source of depth data on its own. To address both issues, we added a new sonar subsystem to get depth data directly, with a new stereo camera that was much easier to calibrate as backup. Our sonar subsystem was developed as a ROS library which can be imported and called by any of the other subsystems. The sonar is controlled through a custom controller which can handle multiple requests from different substems and return the depth of predicted objects.\cite{b3}

\begin{figure}[h!]
\centering
\includegraphics[width=80mm]{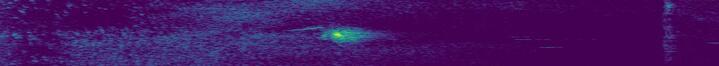}
\caption{Buoy Detected in the Middle of a Sonar Sweep}
\label{fig:method}
\end{figure}

For buoy detection we scan around the angle of the detected object and get the average distance from the detected object. For gate detection, we used OpenCV to implement a "circular similarity" algorithm which ranks each significant curve to how close they are to a perfect circle; the highest ranking curves are then returned as candidates as potential gates. This information is combined with CV camera data to produce beliefs of where objects are locally from the robot. 

\section{Testing Strategy}

\subsection{Test Plan}
The most important lesson from our RoboSub competition last year was that testing is the key to success. Ensuring every component and subsystem is reliable will ensure a successful competition run. Oogway's testing suite comes from a combination of simulated and real-world environments. Our major improvement to our previous design philosophy involves constantly testing parts independently to ensure that their often-complicated integration is successful. Our test plan is as follows:
\begin{enumerate}
    \item Test all components that are compatible with simulation testing in a simulation. This ensures that each component works properly and connects to the robot without causing negatively impacting other systems.
    \item Test each component physically on the robot to ensure all components fit properly. We run component specific tests in the pool and in the sink when possible.
    \item Once all components work separately, test the integration of components in the pool to mimic a real competition environment. Pool testing allowed us to test the robot on completion tasks and ensured we could complete each task reliably.
\end{enumerate}

By following this strategy we made sure that every component was individually reliable and the robot was able to complete all tasks

\subsection{Simulated Testing}

\subsubsection{Acoustics}
While developing the algorithm underlying the acoustics stack we created a simulated acoustics program. This involved simulating a pinger source and calculating the induced pressure waves that would propagate. This, combined with encoded hydrophone geometries and artificial noise provided realistic data to test the algorithm with.

\subsubsection{Software}
We utilized our custom sim environment in CoppeliaSim which includes models of Oogway, the gate, buoys, and the octagon to test controls and task algorithms. Before empirical testing we ran each of our \textit{core tasks} in the sim to ensure our code completed the task correctly.

\subsubsection{Computer Vision}
To test the CV algorithm with non-pool data, we created a simulation in Unity to generate training images. These synthetic images were quickly generated and included varying orientations of the glyphs, multiple positions of the camera, and even accounted for the murkiness of the water in the TRANSDEC pool. Annotations were automatically generated alongside the synthetic images, generating thousands of labeled images extremely quickly. These images strategically supplemented our existing model with data that provided a more uniformly distributed data model.

\subsection{Empirical Testing}

\subsubsection{Mechanical}
All water-facing parts on the robot underwent extensive functionality and water-resistance testing. This often involved long-duration submersion tests in a sink. Once passed, the testing would escalate to a diving pool. This comprehensive testing approach guarantees all vital components will remain watertight during operation. 

\subsubsection{Electrical}
When each component was integrated, we ran them for extended periods of time. This helped to test long-term performance and caught many wiring and power distribution issues. All subcon and epoxy connections undergo a 24 hour submersion water testing. Electrical components such as the cameras and sonar also went thought these same tests. Once all of the components were tested individually, we performed full system integration tests to ensure software could use all of the electrical data.

\subsubsection{Acoustics}
After our algorithm correctly located the pinger in simulation we performed the same test in the pool with a physical pinger and the hydrophones on the robot. We ran the robot to move towards the pinger, making sure we could reach the location and surface properly.

\subsubsection{Software}
Controls is tested through a custom GUI system which allows for robot control through both joystick inputs and desired setpoints. Using setpoints, we observed how the robot moved and adjusted the PID constants accordingly.\cite{b4} For testing individual tasks, we set up objects in a diving pool to mimic a competition experience.

\subsubsection{CV}
To test the object detection algorithm, we set up buoys in a lane pool and manually moved the robot around to test all vision angles. We also tested the same system integrated with the sonar, matching the received sonar data with the CV data. We then tested the robot could move to and hit the two correct abydos symbols.

\section*{Acknowledgment}
Duke Robotics Club is primarily sponsored by the Duke University’s Pratt School of Engineering. We are grateful to the Ali Stocks and the Foundry staff for housing and supporting us. We are also indebted to Director of Undergraduate Student Affairs Jennifer Ganley for helping us plan events, our advisors Professor Michael Zavlanos and Dr. Boyuan Chen for technical assistance, and the Engineering Alumni Council, for overall club advice. Furthermore, we owe our success to our long-time sponsors: The Lord Foundation, Duke Student Government, General Motors, and SolidWorks. Finally, we thank RoboNation, whose commitment to RoboSub empowers student engineers like ourselves to explore our passion for robotics.

\onecolumn
\clearpage

\section*{Appendix A: Component List}

\begin{table}[H]
\centering
\begin{tblr}{
  width = \linewidth,
  colspec = {Q[198]Q[96]Q[185]Q[137]Q[135]Q[69]Q[115]},
}
\hline
\textbf{Component}               & \textbf{Vendor} & \textbf{Model/Type}          & \textbf{Specs}          & \textbf{Custom/Purchased} & \textbf{Cost} & \textbf{Year of Purchase} \\ \hline
Buoyancy Control                 & Blue Robotics   & Subsea Buoyancy Foam         & R-3312                  & Purchased                 & 200.00        & 2023                     \\
Frame                            &                 & Aluminum                     & 821                     & Custom                    & 2,000.00      & 2023                     \\
Waterproof Housing               &                 & Polycarbonate                & -                       & Custom                    & 345.00        & 2023                     \\
Waterproof Casing                & Blue Robotics   & Watertight Enclosure         & 4" Cast Acrylic Plastic & Purchased                 & 124.00        & 2023                     \\
Waterproof Connectors            & MacArtney       & Subconn/Seaconn Connectors   & -                       & Purchased                 & 1600.00       & 2019                     \\
Waterproof Connectors            & Blue Robotics   & WetLink Penetrators          & 4.5MM-HC                & Purchased                 & 400.00        & 2023                     \\
Propulsion                       & Blue Robotics   & T200 Thruster                & -                       & Purchased                 & 1,790.00      & 2023                     \\
Power System                     & Turnigy         & 16000mAh                     & 4S 12C                  & Purchased                 & 210.00        & 2023                     \\
Motor Controls                   & Blue Robotics   & Basic ESC                    & -~                      & Purchased                 & 300.00        & 2023                     \\
High Level Control               & Arduino         & Arduino Nano Every           & -                       & Purchased                 & 13.50         & 2023                     \\
CPU                              & ASUS            & MiniPC PN80                  & -                       & Purchased                 & 1,300.00      & 2022                     \\
Internal Comm Network            & TP-Link         & TL-SG1005P V2                & -                       & Purchased                 & 40.00         & 2023                     \\
External Comm Interface          & NetGear         & Nighthawk R7000              & -                       & Purchased                 & 143.00        & 2019                     \\
Intertial Measurement
Unit (IMU) & VectorNav       & VN-100 Rugged IMU            & -                       & Purchased                 & 1,335.00      & 2020                     \\
Doppler Velocity Logger
(DVL)    & Teledyne        & Pathfinder 600               & -                       & Purchased                 & 10,000.00              & 2023                     \\
Camera(s)                        & Luxonis         & OpenCV AI Kit: OAK D-PoE     & -                       & Purchased                 & 400.00           & 2023                     \\
Hydrophones                      & Aquarian        & H1a Hydrophones              & -                       & Custom                    & 600.00        & 2021                     \\ \hline
Vision~ ~                        & OpenCV          &                              &                         &                           &               &                          \\
Localization Mapping         &                 & Extended Kalman Filter, SLAM &                         & Custom                    &               &                          \\
Autonomy                         &                 & Custom task planner          &                         & Custom                    &               &                          \\ \hline
Open-Source Software             &                 & Docker and ROS               &                         &                           &               &                          \\
Programming Language(s)          &                 & Python, C++, Lua             &                         &                           &               &     
\end{tblr}
\end{table}

\clearpage
\twocolumn

\section{Appendix B: Test Plan Results}

\subsection{Mechanical}
Mechanical testing is arguably the most critical aspect of verifying Oogway's performance. A failure here would disable all other subsystems. As described previously, water-facing components were put through multiple rounds of testing.

The waterproof seal on the capsule for the electrical stack was initially not able to withstand 24 hours submerged in the lab sink. Indication of moisture was detected after the duration. As a result, the attachment mechanism was refined to include a clamping bolt-nut configuration rather than a threaded insert in the capsule. This configuration survived the 24 hours in the sink and performed well in the diving pool. However, only after multiple tests in this diving pool did the seal on the capsule fail. We concluded that this was due to insufficient clamping force on the middle portion of the O-Ring, and a rope attachment was designed to concentrate force there. This design has passed all testing since.

\subsection{Electrical}
Some electrical components did fail our testing procedure. A notable example was an internal relay. This relay was tested long term by having it enabled for over an hour. This constant current draw failed the Arduino it was connected to. The circuit was then redesigned to have an alternative power source and it was converted from normally-open to normally-closed, the typical working condition. Additionally, a software implementation was put in place to track this relay's circuit.

Short circuit tests successfully protected all components. This worked for both the computer and thruster power systems (the two main reference voltages onboard).

\subsection{Software + CV}
The majority of the individual software components passed the component testing without many problems. The CV object detection worked in the pool on the first attempt. After some initial data collection, the sonar correctly identified the buoys and the gate with no additional testing.

However, the integration between CV and sonar required many hours of testing in the pool. We found that the angles of the CV camera and sonar did not always match up so the sonar would scan empty pool space. Additionally, we found that the sonar has a minimum range of about a meter so the robot would run through the buoy without stopping and backing up.

With many pool tests of effort we were eventually successful in locating the buoy and hitting the correct symbols.

\end{document}